%% file: emnlp2022.tex
\pdfoutput=1

\documentclass[11pt]{article}

\usepackage{EMNLP2022}

\usepackage{times}
\usepackage{latexsym}

\usepackage[T1]{fontenc}

\usepackage[utf8]{inputenc}

\usepackage{microtype}

\usepackage{inconsolata}

\usepackage{times}
\usepackage{latexsym}
\usepackage{amsthm,amsmath,amssymb}
\usepackage{mathrsfs}
\usepackage{bm}
\usepackage{booktabs}
\usepackage{color}
\usepackage{graphicx}
\usepackage{multirow}

\usepackage{float}
\usepackage{url}

\newcommand{\squishlist}{
 \begin{list}{$\bullet$}
  { \setlength{\itemsep}{0pt}
     \setlength{\parsep}{3pt}
     \setlength{\topsep}{3pt}
     \setlength{\partopsep}{0pt}
     \setlength{\leftmargin}{1.5em}
     \setlength{\labelwidth}{1em}
     \setlength{\labelsep}{0.5em} } }

\newcommand{\squishend}{
  \end{list}  }

\title{Unsupervised Opinion Summarisation in the Wasserstein Space}

\author{Jiayu Song$^{1}$, Iman Munire Bilal$^{2,3}$, Adam Tsakalidis$^{1,3}$, Rob Procter$^{2,3}$, Maria Liakata$^{1,2,3}$ \\
        $^1$ Queen Mary University of London, London, UK \\ 
        $^2$ University of Warwick, Coventry, UK\\ 
        $^3$ The Alan Turing Institute, London, UK\\
        \texttt{\{jiayu.song,a.tsakalidis,m.liakata\}@qmul.ac.uk}\\}

\begin{document}
\maketitle
\input{1abstract.tex}

\input{2introduction}

\input{3related_work}

\input{4model}

\input{5experiment}
\section{Acknowledgements}
This work was supported by a UKRI/EPSRC Turing AI Fellowship to Maria Liakata (grant no. EP/V030302/1) and The Alan Turing Institute (grant no. EP/N510129/1) through project funding and its Enrichment PhD Scheme.
We are grateful to our reviewers and thank them for their support. We would also like to thank our journalist collaborators for their work on conducting the human evaluation of the system summaries.
\input{7limitation}


\input{anthology.bbl}
\bibliography{anthology,custom}
\bibliographystyle{acl_natbib}

\appendix

\input{6appendix}

\end{document}

%% file: 1abstract.tex
\begin{abstract}



Opinion summarisation synthesises opinions expressed in a group of documents discussing the same topic to produce a single summary. 
Recent work has looked at opinion summarisation of clusters of social media posts. Such posts are noisy and have unpredictable structure, posing additional challenges for the construction of the summary distribution and the preservation of meaning compared to online reviews, which has been so far the focus of opinion summarisation.
To address these challenges we present \textit{WassOS}, an unsupervised abstractive summarization model which makes use of the Wasserstein distance. 
A Variational Autoencoder is used to get the distribution of documents/posts, and the distributions are disentangled into separate semantic and syntactic spaces. The summary distribution is obtained using the Wasserstein barycenter of the semantic and syntactic distributions. A latent variable sampled from the summary distribution is fed into a GRU decoder with a transformer layer to produce the final summary. Our experiments on multiple datasets including Twitter clusters, Reddit threads, and reviews show that WassOS almost always outperforms the state-of-the-art on ROUGE metrics and consistently produces the best summaries with respect to meaning preservation according to human evaluations.


\end{abstract}

%% file: 2introduction.tex
\section{Introduction}

The growth of online platforms has encouraged people to share their opinions, such as product reviews on online shopping platforms (e.g., Amazon) and responses to events posted on social media  (e.g., Twitter). Summarising users' opinions over particular topics on such platforms is crucial for decision-making and helping online users find relevant information of interest \cite{rashid2002getting,fan2019graph}. Specifically multi-document opinion summarisation aims at automatically summarising multiple opinions on the same topic \cite{moussa2018survey}. The bulk of work in this area uses unsupervised summarisation methods.

\vspace{.08cm}
\noindent\textbf{Datasets/Domains}. Most work on unsupervised abstractive opinion summarisation focuses on reviews (e.g., Amazon, Yelp) \cite{wang-ling-2016-neural,chu2019meansum,bravzinskas2020unsupervised,amplayo2020unsupervised,elsahar2020self}. However, it is also important to capture user opinions in online discussions over specific events or topics on popular social media platforms such as Twitter \cite{Template-based} and Reddit,
where the text structure and content is very different and often much noisier compared to review-based corpora (see some examples in Appendix~\ref{apen-examples-twitter} and \ref{apen-examples-review-summaries}). 

\vspace{.08cm}
\noindent\textbf{Summary Representation}. A main focus of unsupervised abstractive summarisation is the creation of a meaningful summary representation.
MeanSum \cite{chu2019meansum} used a text autoencoder to construct summary latent variables by aggregating document latent variables. Subsequent research \cite{bravzinskas2020unsupervised, iso2021convex} adopted a variational autoencoder (VAE)  
\cite{kingma2013auto}, which can capture global properties of a set of documents (e.g., topic). 
As a VAE constructs the distribution of a document, including both semantic and syntactic information, the main meaning may be lost when latent variables sampled from the document distributions are directly aggregated; thus we need methods that can cater for the potential effect of syntactic information, and distinguish between syntax and semantics, especially in documents with unpredictable structure.
However, previous work has not considered syntactic and semantic information separately \cite{bravzinskas2020unsupervised, iso2021convex}. 
Another important consideration is the relative weights of documents within a summary vs obtaining an average~\cite{chu2019meansum, bravzinskas2020unsupervised, iso2021convex}. We mitigate the potential effect of syntactic information on the acquisition of semantic information through a disentangled method. We combine the disentanglement into separate syntactic spaces from \cite{bao2019generating} with the Wasserstein distance and Wasserstein loss to obtain the summary distribution. Our experiments with different settings and datasets prove the validity of this strategy.  \\
\noindent Specifically our work makes the following contributions:

\squishlist
\item We are the first to address multi-document unsupervised opinion summarisation from noisy social media data;

\item we provide a novel opinion summarisation method (\textit{``WassOS''})\footnote{\url{https://github.com/Maria-Liakata-NLP-Group/WassOS}} based on VAE and the Wasserstein barycenter: we disentangle the document distributions into separate semantic and syntactic spaces \cite{bao2019generating}. We introduce these distributions into the Wasserstein space and construct the summary distribution using the Wasserstein barycenter \cite{agueh2011barycenters}. 
This strategy can reduce the mutual interference of semantic and syntactic information, and identify the representative summary distribution from multiple noisy documents;

\item we compare our method's performance with established state-of-the-art (SOTA) unsupervised abstractive summarisation methods on clusters of posts on Twitter, Reddit threads and online reviews;

\item we provide both quantitative evaluation through standard summarisation metrics as well as qualitative evaluation of generated summaries. Our results show that our approach outperforms the SOTA on most metrics and datasets while also showing the best performance on meaning preservation during human evaluation.

\squishend

%% file: 3related_work.tex
\section{Related Work}


\noindent\textbf{Opinion summarization.}
The goal of  opinion summarization is to automatically summarize multiple opinions related to the same topic \cite{moussa2018survey}. The most commonly used datasets consist of reviews~\cite{wang-ling-2016-neural, chu2019meansum, amplayo2020unsupervised, bravzinskas2020unsupervised, iso2021convex}, which assess a product from different aspects and have relatively fixed text structure. On the basis of such datasets, MeanSum \cite{chu2019meansum} uses unsupervised methods to generate abstractive summaries. It uses a text autoencoder to encode each review, and averages the latent variables of each review to get the latent variable of the summary. Subsequently, several works have focussed on obtaining a meaningful summary distribution for this task. \citet{bowman2015generating} and \citet{bravzinskas2020unsupervised} use a variational autoencoder (VAE) \cite{kingma2013auto} to explicitly capture global properties of a set of documents (e.g., topic) in a continuous latent variable. They average these document latent variables to get the summary latent variable and capture the overall opinion. \citet{iso2021convex} argue that input documents should not be treated equally, allowing their model (`COOP') to ignore some opinions or content via the use of different weights for different input documents.
Social media posts, such as those on Twitter, Reddit, and news (i.e. CNN/Daily mail corpus (CNN/DM)\cite{hermann2015teaching}) also express users' opinion. Such datasets are profoundly unstructured and noisy, using casual language~\cite{rao2015model, moussa2018survey}. Recent work on opinion summarisation has considered social media posts using a template-based supervised approach \cite{Template-based}. However the mutual interference of semantic and syntactic information has not been considered. 
Our work explores an effective model for unsupervised opinion summarisation from both social media posts and online reviews, while disentangling syntax from semantics. \\
\vspace{-0.5cm}

\noindent \textbf{Wasserstein distance}.\label{wass}
In most work on generative learning (e.g., text or image generation), it is necessary to calculate the distance between the simulated and the real data distribution. Work from text summarization  \cite{choi2019vae,bravzinskas2020unsupervised} and sentence generation \cite{bowman2015generating}, which uses a VAE, adopts the KL (Kullback–Leibler) divergence, whereas Generative Adversarial Networks (GAN) \cite{goodfellow2014generative} use the JS (Jensen–Shannon) divergence for this purpose, since GANs face issues related to mode collapse caused by the asymmetry of KL divergence. 
However, when there is no overlap between the real and generated distributions, or overlap is negligible, then the corresponding JS or KL distance values can be a constant, leading to the problem of a vanishing gradient. Here we avoid these issues by leveraging the Wasserstein distance to calculate the distance between different document distributions \cite{xu2018distilled, chen2019improving}.

%% file: 4model.tex
\section{Methodology} 
\textbf{Task}. Given a set of documents (here, social media posts or product reviews) on the same topic, the aim is to summarise opinions expressed in them. This section describes our multi-document abstractive summarisation approach which combines a disentangled VAE space with the Wasserstein distance. 

\subsection{Architecture Overview}\label{architecture_overview}
We build our framework on the basis of the Variational Auto-Encoder (VAE, \S\ref{VAE_sem_syn_encode}), which can obtain latent representations from a set of documents both at the level of the individual document and the group
\cite{bravzinskas2020unsupervised}. 
To preserve the meaning of the documents and reduce the impact of noise and purely syntactic information, we disentangle the document representation into (a) semantic and (b) syntactic spaces \cite{bao2019generating} and construct the summary distribution from both.

\begin{figure*}
\includegraphics[width=0.95\textwidth]{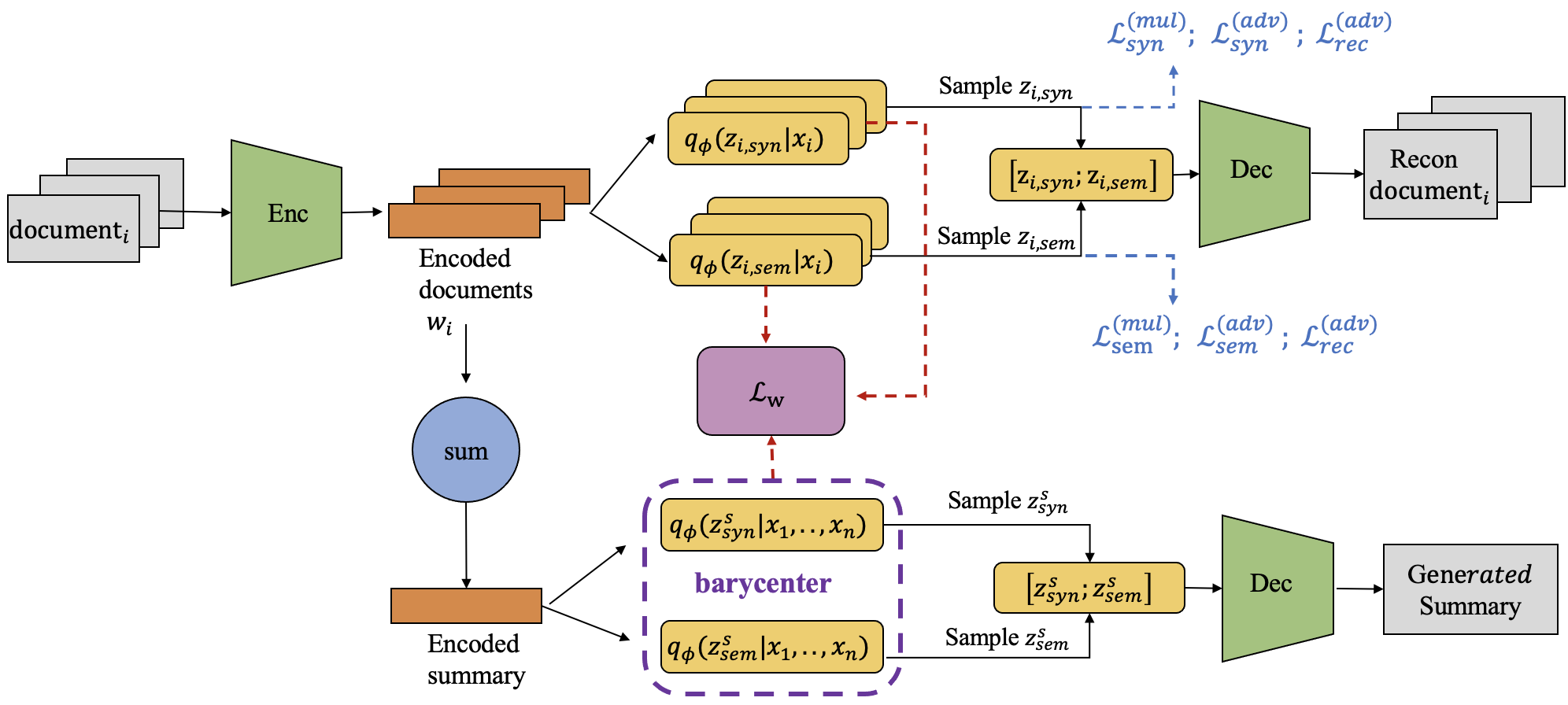}
\caption{Overview of WassOS. The red dashed arrows are Wasserstein losses embedded in the Wasserstein barycenter formula. The blue dashed arrows are multi-task and adversarial losses for disentangling the semantic and syntactic spaces. The figure shows the first strategy to construct the syntactic summary distribution, where the summary latent variable is sampled from the syntactic and semantic barycenters of the document distributions.} 
\label{fig:overview}
\end{figure*}

Unlike earlier work~\cite{chu2019meansum,bravzinskas2020unsupervised,iso2021convex} we construct the summary distribution as the barycenter (the centre of probability mass) of the syntactic and semantic document distributions (see Figure~\ref{fig:overview}). Moreover, to counterbalance the effect of the vanishing gradient resulting from use in the loss function of distance metrics such as KL and JS, we are the first to employ the Wasserstein distance and the corresponding Wasserstein barycenter formula in the context of summarisation (\S\ref{summary_com}). 

Figure~\ref{fig:overview} shows the overall model structure.
$\bold{X} = {\left\{ {x_1},..., {x_i}, ..., {x_n}\right\}}$ denotes a group of documents to be summarised. 
The model consists of three main components:\\ (1) a VAE-encoder (\S\ref{VAE_sem_syn_encode}) that learns distributions for each document ${x_i}$ in separate semantic and syntactic spaces \cite{bao2019generating}, samples the corresponding latent variables ${z_{i, sem}}$ and ${z_{i, syn}}$ and gets the document latent variables ${z_i}$ by combining ${z_{i, sem}}$ and ${z_{i, syn}}$;\\
(2) a summarization component (\S\ref{summary_com}) that learns to construct the syntactic and semantic summary distributions, from which it samples the corresponding latent variables which are concatenated to give the summary latent variable $z^s$. 
The summary semantic distribution $v^s_{sem}$ is the Wasserstein barycenter of all document semantic distributions $v_{i,sem}$ while we examine two different strategies for obtaining the summary syntactic distribution $v^s_{syn}$.\\
(3) Finally, the decoder (\S\ref{decoder_component}) generates the summary by combining an auto-regressive GRU decoder as in \citet{bravzinskas2020unsupervised} with a transformer layer with pre-trained BERT parameters, to guide the generation with syntactic information already encoded in BERT \cite{jiang2020transformer,fang2021transformer}.
We input the summary latent variable ${z^s}$ into the transformer layer, and the output of the transformer is concatenated with the previous state of the GRU decoder \cite{cho2014learning} as input at every decoder step.\vspace{-.15cm}

\subsection{Document Reconstruction through VAE}\label{VAE_sem_syn_encode}
\textbf{Variational Auto-Encoder (VAE)}. We use a VAE to encode a group of documents, disentangle it into semantic and syntactic spaces and sample the corresponding latent variables. Given a group of documents $\{{x}_1,$ ...$, {x}_n\}$, a VAE model will parameterize an approximate posterior distribution ${q_{\phi}(z_i|x_{i})}$ (a diagonal Gaussian) 
\cite{bowman2015generating}. We encode the documents with a GRU encoder as in \citet{bravzinskas2020unsupervised} to get the representation $h_{i}$ of each document. To compute the parameters of the approximate posterior ${q_{\phi}(z_{i}|x_{i})}$=${N({z_{i}};{\mu_{\phi}(x_{i})},I\delta_{\phi}(x_{i}))}$, we linearly project the document representations -- i.e., we use the affine projections to get the Gaussian's parameters: 
\begin{equation}
    \begin{aligned}
    {\mu_{\phi}(x_{i})}={{L}{h_{i}}+{b_{L}}} \\
    {\log{\delta_{\phi}(x_{i})}}={{G}{h_{i}}+{b_{G}}}.
    \end{aligned} \label{affine}
\end{equation}
Then the VAE follows an objective that encourages the model to keep its posterior distributions close to a prior $p({z_i})$, generally a standard Gaussian distribution ($\mu=\overrightarrow{0}$, $\sigma=\overrightarrow{1}$) \cite{bowman2015generating}. This objective is to maximise its lower bound:
\begin{equation}
    \begin{aligned}
    {L(\theta;x_i)} = {-KL({q_{\phi}(z_{i}|x_{i})||p(z_{i})})}\\
    +{\mathbb{E}_{q_{\phi}(z_{i}|x_{i})}\lbrack {\log{p_{\theta}(x_{i}|z_{i})}} \rbrack}\\
    \leq{\log{p(x_{i})}}. \label{vae_loss}
    \end{aligned}
\end{equation}
To capture the opinion expressed in multiple documents, we disentangle the corresponding latent variables into two types -- semantic $z_{i,sem}$ and syntactic $z_{i,syn}$ following \citet{bao2019generating}. In this way, the model can capture semantic and syntactic information separately and reduce their interference. As in \citet{bao2019generating}, Eq.~\ref{vae_loss} becomes:
\begin{equation*}
    \begin{aligned}
    {L(\theta;x_i)} = {-KL({q_{\phi}(z_{i,sem}|x_{i})||p(z_{i,sem})})}\\
    {-KL({q_{\phi}(z_{i,syn}|x_{i})||p(z_{i,syn})})}\\
    +{\mathbb{E}_{{q_{\phi}(z_{i,sem}|x_{i})}{q_{\phi}(z_{i,syn}|x_{i})}}\lbrack {\log{p_{\theta}(x_{i}|z_{i,sem},z_{i,syn})}} \rbrack}\\
    \leq{\log{p(x_{i})}}. \label{vae_sem_syn_loss}
    \end{aligned}
\end{equation*}
In the description that follows, we denote ${q_{\phi}(z_{i,sem}|x_{i})}$ and ${q_{\phi}(z_{i,syn}|x_{i})}$ as $v_{i,sem}$ and $v_{i,syn}$ respectively.
We adopt the multi-task, adversarial losses and adversarial reconstruction losses of the DSS-VAE model \cite{bao2019generating}. We assume $z_{i,sem}$ to predict the bag-of-words (BoW) distribution of a sentence, whereas $z_{i,syn}$ is used to predict the tokens in a linearized parse tree sequence of the sentence separately. Their losses are respectively defined as:
\begin{equation*}
    \begin{aligned}
    {\mathcal{L}_{sem}^{(mul)}}
     = {-\sum_{w\in V} {t_w}{\log p(w|z_{i,sem})}}
    \label{mul_sem_loss}
    \end{aligned}
\end{equation*}\vspace{-.3cm}
\begin{equation*}
    \begin{aligned}
    {\mathcal{L}_{syn}^{(mul)}}
     = {-\sum_{j=1}^{n}}{\log p({s}_{j}|{s}_{1}...{s}_{j-1},z_{i,syn})},
    \label{mul_syn_loss}
    \end{aligned}
\end{equation*}
where $\bm{t}$ is the ground truth distribution of the sentence, $p(w|z_{i,sem})$ is the predicted distribution and  $s_j$ is a token in the linearized parse tree.

The adversarial loss in DSS-VAE further helps the model to separate semantic and syntactic information. It uses $z_{sem}$ to predict token sequences, but predicts the bag-of-words (BoW) distribution based on $z_{syn}$.
The VAE is trained to `fool' the adversarial loss by minimizing the following losses: \vspace{-.25cm}
\begin{equation*}\vspace{-.25cm}
    \begin{aligned}
    {\mathcal{L}_{sem}^{(adv)}}
     = {\sum_{w\in V} {t_w}{\log p(w|z_{i,syn})}}
    \label{adv_sem_loss}
    \end{aligned}
\end{equation*}
\begin{equation*}
    \begin{aligned}
    {\mathcal{L}_{syn}^{(adv)}}
     = {\sum_{j=1}^{n}}{\log p({s}_{j}|{s}_{1}...{s}_{j-1},z_{i,sem})}
    \label{adv_syn_loss}
    \end{aligned}
\end{equation*}
Furthermore, DSS-VAE proposes adversarial reconstruction loss to discourage the sentence being predicted by a single latent variable $z_{i,sem}$ or $z_{i,syn}$. The loss is imposed by minimizing:
\begin{equation}
    \begin{aligned}
    {\mathcal{L}_{rec}^{(adv)}}(z_t)
     = {\sum_{i=1}^{M}}{\log p_{rec}({x}_{i}|{x}_{<i},z_{t})},
    \label{adversarial_reconstruction_loss}
    \end{aligned}
\end{equation}
where ${M}$ is the length of the sentence, and ${z_t}$ is $z_{i,syn}$ or $z_{i,sem}$.

\subsection{Summarization Component}\label{summary_com}
This is the core component for constructing the summary distribution. After obtaining the distribution of each document in a group, we seek to obtain the distribution of a hypothetical summary of the group of documents. 
Our intuition is to directly initialize a summary distribution that has the smallest distance from a group of document distributions. In this way, we impose a higher semantic similarity between the generated summary and the group of documents and increase the chance that the generated summary can capture the opinions expressed in the group of documents. We set the following minimization problem as our training objective:\vspace{-.25cm} 
\begin{equation}\vspace{-.25cm}
    \begin{aligned}
    {\inf_{v^s}}{\sum_{i=1}^{n}{\lambda_{i}{D}({v_i},{v^s})}}, \label{distribution_distance}
    \end{aligned}
\end{equation}
where $n$ is the number of documents, ${D}({v_i},{v^s})$ is the distance between a document distribution ${v_i}$ and the summary distribution $v^s$, and $\lambda_{i} = {f(z_{i})}$ is the weight of the distance between the summary and each of the document distributions.
$f$ is implemented as a feed forward network. 
Considering the advantages of the Wasserstein distance (see \S\ref{architecture_overview}), we introduce the document distributions into the Wasserstein space and use the Wasserstein distance as $D$ in formula~\ref{distribution_distance}. This allows us to calculate 
the Wasserstein barycenter $v^s$ of the document distributions. The barycenter provides the centre of probability mass between distributions. \vspace{-.10cm}

\paragraph{Wasserstein Barycenter in Gaussian} \citet{agueh2011barycenters} propose the definition of a barycenter in the Wasserstein space. In analogy to the Euclidean case, where the barycenter is calculated on the basis of formula~\ref{distribution_distance}  with $D$ being the squared Euclidean distance, they replace the squared Euclidean distance with the squared 2-Wasserstein distance:
defined as:
\begin{equation}
    \begin{aligned}
    {W^{2}_{2}(P,Q)}={||\mu_{1}-\mu_{2}||}^{2} + {{B}^{2}(\Sigma_{1},\Sigma_{2})}, 
    \end{aligned} \label{gaussian}
\end{equation}
where ${B}^{2}(\Sigma_{1},\Sigma_{2})$ is:
\begin{equation*}
    \begin{aligned}
    {tr(\Sigma_{1}) + tr(\Sigma_{2})} - 
    {2tr {\lbrack{\Sigma_{1}^{1/2}} {\Sigma_{2}} {\Sigma_{1}^{1/2}})^{1/2} \rbrack}}
    \end{aligned}
\end{equation*}
They then minimize\vspace{-.15cm}:
\begin{equation}\vspace{-.15cm}
    \begin{aligned}
    {\inf_{v}}{\sum_{i=1}^{p}{\lambda_{i}{W^{2}_{2}}({v_i},{v})}}, 
    \label{wass-barycenter}
    \end{aligned}
\end{equation}
where $v_i$ and $v$ are probability distributions,
$\lambda_i$'s are positive weights summing to 1 and ${W^{2}_{2}}$ denotes the squared 2-Wasserstein distance. 

Since the distributions assumed in VAE \cite{kingma2013auto} are Gaussian, it is important to know whether the barycenter exists in this case, and the corresponding specific Wasserstein distance formula. \citet{agueh2011barycenters} proved the existence and uniqueness of the barycenter in problem ~\ref{wass-barycenter} in the Gaussian case, and provided an explicit formula.  
However, this formula is only applicable when $\mu$ is 0, that is $Gaussian(0,\sigma_{i}^{2})$. A proof by \citet{delon2020wasserstein} demonstrates that if the covariances $\Sigma_{i}$ are all positive definite, then the barycenter exists for Gaussian distributions.
The above studies provide the theoretical support for our model, which obtains the Wasserstein barycenter as the summary distribution under the assumptions of a VAE \cite{kingma2013auto}. \vspace{-.15cm}

\paragraph{Wasserstein distance in Gaussian}
Next, we consider the calculation of the Wasserstein distance under the assumptions of a VAE.
\citet{kingma2013auto} provide the theory for an Auto-Encoding Variational Bayes. They assume that all prior distributions are Gaussian, and that true posteriors are approximately Gaussian with an approximately diagonal covariance. In this case, they let the variational approximate posteriors be multivariate Gaussian with a diagonal covariance structure:
\begin{equation*}
    \begin{aligned}
    {\log{q}_{\phi}({z|x})}={\log{\mathcal{N}}(
    {z};{\mu_{z}},{\delta_{z}^{2}I})}
    \end{aligned}
\end{equation*}
Thus the Wasserstein distance (Eq.~\ref{wass_guassian}) can be derived in the Gaussian case from Eq.~\ref{gaussian}, where the two Gaussian distributions are multivariate Gaussians with a diagonal covariance:
\begin{equation}
    \begin{aligned}
    {W^{2}_{2}}={\sum_{j=1}^{J} \lbrack(\mu_{1j}-\mu_{2j})^{2} + \delta_{1j}^{2} + \delta_{2j}^{2} -2(\delta_{1j}^{2}.\delta_{2j}^{2})^{\frac{1}{2}}\rbrack ,}
    \end{aligned}\label{wass_guassian}
\end{equation}
where $J$ is the dimensionality, and $\mu_{j}$, $\delta_{j}$ denote the {j}-th element of ${\mu}$ and ${\delta}$, respectively. 

Based on the above theory, we can assume that there is a posterior distribution of a summary of documents, expressed as the barycenter of the document distributions, which is a multivariate Gaussian with a diagonal covariance structure:
\vspace{-.3cm}
\begin{equation*}
    \begin{aligned}
    {\log{q}_{\phi}({{z^s}|x_{1},...,x_{n}})}={\log{\mathcal{N}}({z^s};{\mu_{{z^s}}},{\delta_{{z^s}}^{2}I})}
    \end{aligned}
\end{equation*}
Specifically, we linearly project the summary representation ${h_s}$ to get the approximate posterior $v^{s}$=${q_{\phi}(z^{s}|x_{1},...,x_{n})}$=${N({z^s};{\mu_{\phi}(h_{s})},I\delta_{\phi}(h_{s}))}$ of the summary, which is the same process as getting the document posterior distribution (Eq.~\ref{affine},\S\ref{VAE_sem_syn_encode}). 
    ${h_s}={{w_1}{h_1}+ ...+ {w_n}{h_n}}$,
where $w_{i}=f(h_i)$ is the weight for each document representation $h_{i}$.

We use Eq.~\ref{wass_guassian} to calculate the Wasserstein distance between the document distributions ${v_i}$ and the assumed summary distribution ${v^s}$ under the assumption of a VAE. Therefore, the final Wasserstein loss function is: \vspace{-.24cm}
\begin{equation*}\vspace{-.24cm}
    \begin{aligned}
    {\mathcal{L}_{wass}}={\inf_{v^s}}{\sum_{i=1}^{n}{\lambda_{i}{W^{2}_{2}}({v_i},{v^s})}} 
    \end{aligned}
\end{equation*}
where the $\lambda_i$'s are positive weights summing to 1 and $n$ is the number of documents in the group.

As elaborated in \S\ref{VAE_sem_syn_encode}, we disentangle the document distribution into two parts which capture semantic and syntactic information separately. Therefore,
we assume summary distributions ${v^{s}_{sem}}$ and ${v^{s}_{syn}}$ in semantic and syntactic spaces respectively, and obtain the corresponding Wasserstein losses.
\begin{equation*}
    \begin{aligned}
    {\mathcal{L}_{w_{sem}}}={\inf_{v^{s}_{sem}}}{\sum_{i=1}^{n}{\lambda_{i}{W^{2}_{2}}({v_{i,sem}},{v^{s}_{sem}})}} 
    \end{aligned}
\end{equation*}\vspace{-.3cm}
\begin{equation*}
    \begin{aligned}
    {\mathcal{L}_{w_{syn}}}={\inf_{v_{syn}^{s}}}{\sum_{i=1}^{n}{\lambda_{i}{W^{2}_{2}}({v_{i,syn}},{v_{syn}^{s}})}} 
    \end{aligned}
\end{equation*}
We sample $z_{sem}^{s}$ from the summary semantic distribution ${v_{sem}^{s}}$, which is the Wasserstein barycenter of all document semantic distributions $v_{i,sem}$. Considering the potential effect of syntactic information in different datasets (e.g., data from social media with a more cluttered text structure), we consider two strategies to obtain the summary syntactic distribution ${v_{syn}^{s}}$: (a) similarly to the above method for ${v_{sem}^{s}}$; (b) use the affine projection applied to document representations in the syntactic space to project the summary representation $h_{s}$ to the summary syntactic distribution. Then we sample ${z_{syn}^{s}}$ from it. Finally, the latent summary variable ${z^{s}}$ is defined as:

\vspace{-.5cm}
\begin{equation*}
    \begin{aligned}
    {z^{s}} = [z_{syn}^{s};z_{sem}^{s}] \label{double_center}
    \end{aligned}
\end{equation*}\vspace{-.6cm}


\noindent We minimize the final loss function, defined as:
\begin{equation}
    \begin{aligned}
    {L}= \sum_{i=1}^{n} [{-L(\theta;x_i)}+ {\mathcal{L}_{sem}^{(mul)}} + {\mathcal{L}_{syn}^{(mul)}} + \\
    {\mathcal{L}_{sem}^{(adv)}} + {\mathcal{L}_{syn}^{(adv)}} +{\mathcal{L}_{rec}^{(adv)}}]
     +{\mathcal{L}_{w}} \label{final_loss}
    \end{aligned} 
\end{equation}
where ${\mathcal{L}_{w}}$ is the Wasserstein loss. For the first strategy (a) ${\mathcal{L}_{w}}= {\mathcal{L}_{w_{sem}}}+ {\mathcal{L}_{w_{syn}}}$. For the second strategy (b), there is no need to calculate the Wasserstein barycenter in the syntactic space, and therefore  ${\mathcal{L}_{w}}= {\mathcal{L}_{w_{sem}}}$.
\begin{figure}
\centering
\includegraphics[width=.8\linewidth]{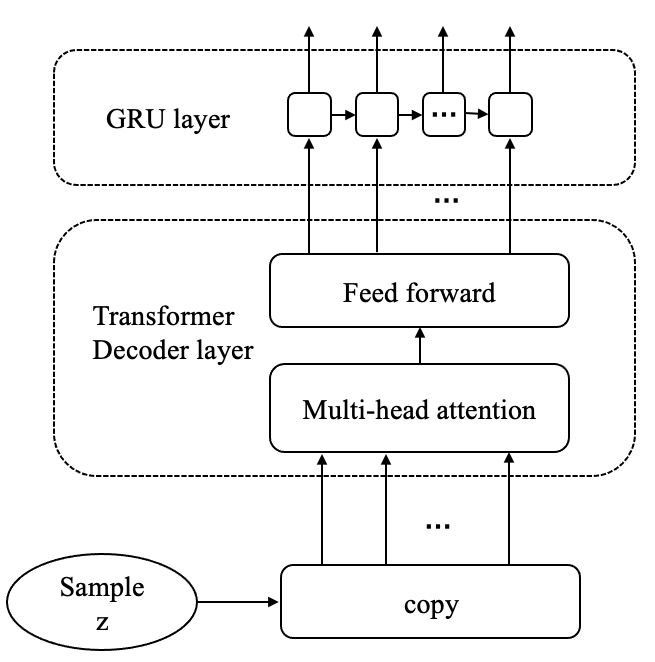}
\caption{The sampled latent variable $z$ is input to the decoder into the transformer layer, and the output of the transformer is concatenated with the input of the GRU. }
\label{decoder}
\end{figure}
\subsection{Decoder component}\label{decoder_component}
In the decoder component, we use an auto-regressive GRU decoder and a pointer-generator network, as in~\cite{bravzinskas2020unsupervised}. In order to make the generated summary more grammatical, we first input the sampled latent variable $z$ into the transformer layer, and then concatenate the output of the transformer to the GRU decoder input at every decoder step, as shown in Figure~\ref{decoder}. The transformer decoder layer contains a multi-head attention layer and a feed forward layer. We load the pre-trained middle layer parameters from BERT \cite{devlin2018bert}, which have been shown to have syntactic features \cite{jawahar2019does}. The same decoder is used for both document reconstruction and summary generation.

%% file: 5experiment.tex
\section{Experiments}
\subsection{Datasets}
We experimented on datasets with different types of content (social media posts, reviews) to allow for a thorough evaluation across different domains:

\noindent\underline{\textbf{Twitter}} \citet
{munire2021evaluation} released 2,214 clusters of tweets on the topics of COVID-19 (2020-2021) and politics (2014-16), manually labeled as being coherent. Each cluster contains $\sim$30 tweets discussing the same sub-topic, posted by different users on the same day. We randomly selected 2,030 clusters for training and 115 for validation of the VAE reconstruction component. We additionaly used 35 clusters for development (GRU) and 34 for overall testing.

\noindent\underline{\textbf{Reddit}} We collected 4,547 Reddit threads from the \texttt{r/COVID19\_support} subreddit, using the PushShift API. We focused on 118 threads with at least 7 comments, to have enough content to perform summarisation. In each thread, we only kept the original post with its comments, ignoring any replies to comments to ensure all content was on topic. 
Finally, we manually selected 40 threads whose posts introduce information pertinent to the topic and do not exceed 70 tokens (similar to the Amazon dataset). Three expert summarisers, native English speakers with a background in journalism were employed to summarise the main story and opinions of each thread, following the same methods used in \cite{Template-based} to create opinion summaries for Twitter. For details regarding the summarisation guidelines see Appendix~\ref{apen-GUI}.
We use these 40 Reddit threads for evaluation purposes only. 

\noindent\underline{\textbf{Amazon}}
\citet{bravzinskas2020unsupervised} released 60 gold summaries for the Amazon product review dataset. 
We follow their work and use 28 products for development and 32 for testing. Furthermore, we use 183,103 products for training the VAE to reconstruct the reviews and 9,639 products for validation -- with 4.6M and 241K reviews, respectively.

\subsection{Models \& Baselines}
We compare our method against existing models for unsupervised abstractive opinion summarisation: 

\smallskip
\noindent\textbf{Copycat} 
\cite{bravzinskas2020unsupervised} relies on continuous latent representations to generate a summary. They use a hierarchical architecture to obtain the distribution of a group of reviews; then, the summary latent variable is chosen by sampling from the distribution of documents within the group.

\smallskip
\noindent\textbf{Coop} \cite{iso2021convex} optimizes the latent vector using the input-output word overlap to overcome the summary vector degeneration. Compared to the averaging strategy in copycat, it calculates the contribution of each review, and has a better performance on review datasets.\smallskip

We also introduce two extractive summarisation baselines that make use of the Wasserstein and Euclidean distance -- \textbf{Medoid (Wass)} and \textbf{Medoid (Eucl)}, respectively --
selecting a single central item (i.e., the `medoid') from a group of documents as the summary. For Medoid (Wass)/Medoid (Eucl), we calculate the Wasserstein/Euclidean distance between each document distribution and the rest and select the document whose distribution is closest to other documents' distributions.\smallskip 

We create two variants of our model to obtain the latent variables of the summary: \textbf{WassOS(T-center)} uses two Wasserstein barycenters (see \S\ref{summary_com}), whereas \textbf{WassOS(O-center)} uses only one Wasserstein barycenter which comes from the summary semantic distribution.


\begin{table}[!h]
\centering
\resizebox{\linewidth}{!}{%
\begin{tabular}{l|lll|lll|lll|}
\cline{2-10} 
&
\multicolumn{3}{|c}{Twitter}
&\multicolumn{3}{|c}{Amazon}
&\multicolumn{3}{|c|}{Reddit}\\
\cline{2-10} 
&R1 & R2 & RL
&R1 & R2 & RL
&R1 & R2 & RL
\\ \hline
Copycat & .305 & .110  & .250
& .319 & .058  & .201
& .206 & .039  & \textbf{.159}
\\ 
Coop & .327 & .135  & .267
& \textbf{.365} & .072  & .212
& .197 & .031  & .137
\\ 
Medoid (Wass) & .264 & .083  & .201
& .288  & .051   & .173
& .164  & .021  & .118
\\
Medoid (Eucl) & .270  & .089 & .219
& .309  & .063 & .189
& .173 & .029 & .119
\\
\hline
WassOS (T-center)    & \textbf{.343} & \textbf{.150}  & \textbf{.291} 
& .285 & .058  & .182
& \textbf{.207} & \textbf{.043 } & .153
\\
WassOS (O-center)      & .265 & .102 & .221 
& .330 & \textbf{.090} & \textbf{.218}
& .174 & .030 & .126
\\
\hline
\end{tabular}
}%
\caption{\label{result-all} ROUGE scores on the test sets (best scores shown in \textbf{bold}). The scores of Coop and Copycat on the Amazon dataset are copied from \citet{bravzinskas2020unsupervised} and \citet{iso2021convex}.}
\end{table}

\subsection{Experimental Settings} 

Before the GRU decoder, we add a transformer layer to provide syntactic information to our model. Since the middle layers from BERT \cite{devlin2018bert} are shown to encode syntactic features (see \S\ref{decoder_component}), 
for our transformer layer we load the pre-trained parameters from the 6$^{th}$ layer\footnote{We tried the middle layers from 5th to 7th in turn, and we found that the model shows the best performance with the 6th layer's parameters.} of \texttt{bert-base-uncased}. 
The text and summary latent variables have the same hidden size as \texttt{bert-base-uncased} (768). We use Adam optimizer \cite{kingma2014adam} (learning rate: $5$$\times$$10^{-4}$). During training, we parse each document into the tag sequence with Zpar\footnote{\url{https://www.sutd.edu.sg/cmsresource/faculty/\\yuezhang/zpar.html}} \cite{zhang2011syntactic}, which serves as the ground truth when getting the syntactic information.

\section{Results}

\subsection{Automatic evaluation}
Results on the test sets are shown in Table~\ref{result-all}. ROUGE-1/2/L scores are based on F1 \cite{lin2004rouge}.

WassOS outperforms all competing models on Twitter, offering a relative improvement of 5\%/11\%/9\% (ROUGE-1/2/L, respectively) over the second best-performing model. On Amazon, it trails by .035 (11\%) in ROUGE-1, but outperforms Coop on ROUGE-2 (25\% improvement) and ROUGE-L.
The results of Copycat and WassOS on Reddit are similar.\footnote{Here we use pre-trained parameters from Twitter.} Copycat slightly outperforms WassOS on ROUGE-L (.006), while WassOS is slightly better on ROUGE-1,2 (.001, .004).

WassOS(T-center) performs better on the Twitter clusters and Reddit threads, but WassOS(O-center) outperforms WassOS(T-center) on Amazon. We hypothesise this is caused by the different acquisition of syntactic latent variables, demonstrating  that syntactic information has an influence on the generated summary. This is likely due to the different format between Amazon reviews and the Twitter/Reddit posts: Amazon reviews follow a very similar format, whereas posts on Twitter/Reddit vary greatly in their structure.
We also make a comparison between two extractive methods based on WassOS, which use two different distances to get the medoid in a cluster of documents. Medoid (Eucl) slightly outperforms Medoid (Wass) on these datasets.  They are both outperformed by WassOS by a large margin.

\begin{table}
\centering
\resizebox{.9\linewidth}{!}{%
\begin{tabular}{l|rrr|rrr|}
\cline{2-7}
&\multicolumn{3}{|c}{Amazon}
&\multicolumn{3}{|c|}{Twitter}\\
\hline  Model &R1 & R2 & RL &R1 & R2 & RL \\ 
\hline
WassOS-dis & .251 & .049  & .175 & .320 & .138  & .272 \\ 
WassOS-trans & .258 & .043  & .173  & .276 & .102  & .236\\
WassOS     & \textbf{.330} & \textbf{.090} & \textbf{.218}& \textbf{.343} & \textbf{.150}  & \textbf{.291} \\\hline
\end{tabular}
}%
\caption{\label{ablation-all}Ablation study: ROUGE on Amazon/Twitter. }
\end{table}

\subsection{Ablation}
We performed ablation studies to investigate the importance of the disentangled component (\S\ref{VAE_sem_syn_encode}) and the transformer decoder (\S\ref{decoder_component}). 
We hypothesize that having messy syntactic information will impact the acquisition of the core meaning. Therefore, we disentangle the latent representation into separate semantic and syntactic spaces, and get the semantic and syntactic information separately. To test the contribution of this approach, we remove the disentangled component. Furthermore, we also tested whether the transformer layer provides syntactic guidance when generating the summary.
In particular, we experimented with (a) removing the disentangled part but keeping the transformer decoder (`WassOS-dis') and (b) keeping the disentangled part but removing the transformer decoder (`WassOS-trans'). We conducted experiments with the two models on the Amazon and Twitter datasets. In `WassOS-trans', we use the first strategy (two barycenters) for Twitter and the second strategy (one barycenter from semantic space) for Amazon. As `WassOS-dis' lacks the disentangled component it uses a single barycenter. Our Reddit dataset is small and is used only for evaluation purposes so does not feature in this comparison where we would have to retrain the model with each of the components removed.

Tables~\ref{ablation-all} shows the ROUGE values on the Amazon and Twitter datasets, respectively. The two models fail to compete against WassOS, showing a drop in ROUGE when either component is removed.
Upon manual investigation of the characteristics of the generated summaries, we find that WassOS-dis (which misses the disentanglement component) often produces summaries with confusing semantic information, as opposed to WassOS-trans (see examples in Tables~\ref{ablation-amazon} and \ref{ablation-twitter} in Appendix~\ref{apen-example-ablation-summaries}). However
the summaries generated by WassOS-dis are more fluent than the summaries generated by WassOS-trans. This shows that the pre-trained parameters on BERT in the decoder component provide helpful syntactic features for the generated summary.
Importantly, our findings highlight that using the transformer or disentangled part alone is not enough to generate good summaries and that both components are equally important to model performance.

\begin{table}[!h]
    \centering
    \resizebox{.95\linewidth}{!}{%
    \begin{tabular}{c|ccccc}
         \toprule

         & & Non- & Referential &  & Meaning \\
         &Model & redundancy & Clarity & Fluency & Preservation \\
        \midrule
        \parbox[t]{2mm}{\multirow{3}{*}{\rotatebox[origin=c]{90}{Twitter}}}&
         Copycat & -.137 & -.078 & -.333 & -.142\\ 
         &Coop & \textbf{.338} & -.323 & \textbf{.363} & -.289\\
         &WassOS & -.201 & \textbf{.402} & -.029 & \textbf{.431}\\
         \midrule
         
         \parbox[t]{2mm}{\multirow{3}{*}{\rotatebox[origin=c]{90}{Reddit}}}&
         Copycat & -.064 & -.113 & .039 & .167\\ 
         &Coop & \textbf{.338} & -.157& -.098 & -.882\\
         &WassOS & -.274 & \textbf{.270} & \textbf{.059} & \textbf{.716}\\
         \midrule
         \parbox[t]{2mm}{\multirow{3}{*}{\rotatebox[origin=c]{90}{Amazon}}}&
         Copycat & \textbf{.517} & \textbf{.420} & \textbf{.207} & -.115\\ 
         &Coop & .144 & .057 & .092 & -.103\\
         &WassOS & -.638 & -.477 & -.299 & \textbf{.218}\\
         \bottomrule
         
    \end{tabular}
    }%
    \caption{Best-Worst evaluation (best scores in bold).}
    \label{HE_all}
\end{table}
\normalsize
\subsection{Human evaluation}
Our last part of the evaluation involves human assessments of the quality of generated summaries. Three experienced journalists, whose professional training includes writing summaries of articles, with previous experience in evaluating NLP generated summaries, were hired for this task. For each entry in the test set (29 test products from Amazon, 34 test clusters from Twitter and 40 test threads from Reddit), we grouped the corresponding generated summaries from Copycat, Coop and WassOS in a summary tuple, assessed by the experts using Best-Worst Scaling \cite{louviere1991best,louviere2015best}. The experts were asked to highlight the best and the worst summary in each tuple with respect to these criteria: \textit{Non-redundancy} (NR), \textit{Referential Clarity} (RC), \textit{Fluency} (F) and \textit{Meaning Preservation} (MP). We describe these criteria in Appendix~\ref{apen-HE}



The results of the human evaluation for the three datasets are shown in Tables \ref{HE_all}. In line with \citet{bravzinskas2020unsupervised}, the final scores (per criterion) for each model are computed as the percentage of times the model was chosen as best minus the percentage of times it was chosen as worst. The scores range between -1 (always chosen as worst) and 1 (always best).

WassOS consistently outperforms Copycat and Coop on meaning preservation (see examples in Tables~\ref{amazon_summary result} and \ref{tweet_summary result} in the Appendix) and also performs well on Twitter and Reddit on referential clarity. We investigated the poor performance of WassOS on Amazon with respect to referential clarity by counting the respective number of pronouns on Amazon and Twitter in iteratively selected samples of equal size.
We found that referential relationships in Twitter are relatively simple compared to Amazon (more details can be found in Appendix~\ref{apen-HEA}). 

We hypothesize that WassOS's suboptimal performance on non-redundancy (NR) is partly due to the degeneration caused by beam search \cite{holtzman2019curious}, 
but also the latent syntactic and semantic representations introducing some redundancy to the decoder (compared to WassOS-dis, the summaries generated by WassOS-trans have more repeated words in Tables~\ref{ablation-amazon} and \ref{ablation-twitter} in the Appendix). 
Future work could look at further optimising disentanglement to avoid redundancy.
Copycat and Coop show widely varying performance on different datasets according to the NR, RC and F criteria and are performing much worse on meaning preservation than WassOS. 

\section{Conclusions}
We present an unsupervised multi-document abstractive opinion summarisation model, which captures opinions in a range of different types of online documents including microblogs and product reviews. 
Our model (`WassOS') disentangles syntactic and semantic latent variables to keep the main meaning of the posts, and uses the Wasserstein loss embedded in the Wasserstein barycenter to obtain a latent representation of the summary. WassOS has the best performance on meaning preservation according to human evaluation across all datasets and outperforms state-of-the-art systems on ROUGE metrics for most datasets. Future work can look into improving non-redundancy and referential clarity of the generated summaries.

%% file: 7limitation.tex
\section*{Limitations}


In our work, we focus on summarizing multiple opinions expressed in documents under the assumption that these are related to the same topic. All of the datasets we have performed our experiments on meet this requirement: (a) we have selected `good' (coherent) clusters from Twitter; (b) we have eye-balled and selected threads from Reddit that can be summarised; (c) each cluster of reviews on the Amazon dataset refers to the same product. It is not evident from our work -- and no conclusions should be reached on -- how our model and baselines would perform if no pre-clustering is performed (i.e., if we are trying to summarise noisy (non-coherent) clusters of documents). Another limitation of our work stems from the fact that the document clusters we have worked on have a restricted number of documents, ranging from 8 reviews in Amazon (as in previous work) to no more than 30 posts for Twitter and Reddit: it is unclear how any of our models/baselines would perform on much larger clusters. 
Although we have performed experiments in a variety of datasets with different linguistic characteristics, the list of domains to explore is non-exhaustive; for example, our model may not be suitable for processing long documents -- and has not been tested in a domain with such characteristics. Last but not least, our work has not focused on characterising diverse and/or conflicting opinions about the same topic, if such opinions co-exist within the same cluster. This aspect may be important in real-world applications aiming at summarising and quantifying diverse opinions. 

%% file: 6appendix.tex
\section{Appendix}

\label{sec:appendix}

\subsection{Human Evaluation Criteria} \label{apen-HE}

\squishlist
\item \textbf{Non-redundancy} (NR): a non-redundant summary should contain no duplication, i.e. there should be no overlap of information between its sentences.

\item \textbf{Referential Clarity} (RC): it should be easy to identify who or what the pronouns and noun phrases in the summary are referring to. If a person or other entity is mentioned, it should be clear what their role in the story is. 

\item \textbf{Fluency} (F): sentences in the summary should have no formatting problems, capitalization errors or ungrammatical sentences (e.g., fragments, missing components) that make the text difficult to read. 

\item \textbf{Meaning Preservation} (MP): a summary preserves the meaning if it presents the same entities and identifies the same correct information about them compared to the gold-standard(s).

\squishend


\subsection{Guidelines for the Creation of human summaries from Reddit threads}\label{apen-GUI}

\subsection*{Overview}
You will be shown a succession of complete Reddit threads which you will be asked to summarise. 
Each thread contains a title and between 8-20 posts where the first post is the OP post (original poster) and the subsequent posts are replies to the OP post. Note that the author of the OP post is also the author of the thread title and is often denoted as OP user. For some reddit threads, the OP post is often the continuation of the text in the title.
Depending on the proportion of opinions expressed in the thread, you will be asked to write a short structured summary (between 20-50 words). Each summary should be well-organised, in English, using complete sentences.
Note that the main story is the focus of the thread and it often describes an objective event. An opinion is defined to be a subjective reaction to the main story.

\subsection*{Steps}
\begin{itemize}
\item Is it possible to easily summarise the opinions within the thread?
Choose “Yes” if there are clear opinions expressed in most of the thread which can be easily summarised. (This option will be suitable for most threads presented to you.)
\item Choose "No” if there exist very few (or no) clear opinions expressed in the thread, but a main story can be easily detected and summarised.
\end{itemize}

\subsection*{Summarisation}  
\begin{itemize}
\item If you responded "No" to Step 1, you will be asked to:
Briefly summarise the main story of the thread (\textasciitilde 20 words)
\item If you responded "Yes" to Step 1: 
Briefly summarise the main story of the thread (\textasciitilde 20 words)
Summarise the opinions expressed in the thread and include any evidence mentioned. (\textasciitilde 20 words)
\end{itemize}

\subsection*{Examples}

Example 1: Reddit Thread with clear opinions in most posts (Yes)

\begin{table}[!htbp]
        \scriptsize
     \centering
     \begin{tabular}{p{7.5cm}}
      \toprule
       \multirow{4}{7.5cm}{\textbf{Title}: Who else is in the same boat? Haven't seen my boyfriend for 1 month, probably won't
  for another 2-3 months and I am at my breaking point. Lockdown is lonely.}\\ \\ \\ \\
 \hline
       \textbf{OP Post}: Note- I live in Canada.\\
       \hline
\multirow{3}{7.5cm}{\textbf{Reply 1}: I live in italy. Haven't seen him since the 8th because of corona. Who knows when we will be able to meet. I feel for you. It's awful. It's heartbreaking.} \\ \\ \\ 
 \hline
\multirow{4}{7.5cm}{\textbf{Reply 2}: Same boat. I'm in Spain. I haven't seen him in two weeks and I don't know when I will. 
 I'm used to seeing him almost every day so it's very frustrating. It helps to try and do some activities together. We do workouts together on skype and play online UNO.} \\ \\ \\ \\
 \hline
\multirow{4}{7.5cm}{\textbf{Reply 3}: I am but with a shorter timeline (it’s only been about a week for me) and I feel it 100
 I would have outlined my problems a few times on here but for some reason I’m not allowed to post? regardless I’m in complete solidarity man.}\\ \\ \\ \\
 \hline
 \multirow{3}{7.5cm}{\textbf{Reply 4}: I haven't seen my girlfriend since the last 5 months. She lives in a different state approximately 500miles away from me. So hang in there you'll pull through this.}\\ \\ \\
 \hline
\textbf{Reply 5}: I am. It’s really awful. \\
 \hline
\textbf{Reply 6}: Same here. Lockdown :(. \\
 \hline
\textbf{Reply 7}: It sucks! \\
 \bottomrule       
     \end{tabular}

     \caption{Example of Reddit thread 1}
     \label{tab:example1}
 \end{table}
 \normalsize
 \noindent \textbf{Summary of main story:} Users discuss missing their significant others who don’t live with them during lockdowns in different parts of the world.
 \noindent \textbf{Summary of opinions:} They share their feelings on loneliness and seek to encourage one another.
 \noindent \textbf{Reasoning:} The thread contains opinions and sentiments shared in almost every post, hence the thread can have both its main story as well as its opinions summarised.

 Example 2: Reddit Thread with few or no opinions (No)

 \begin{table}[!htbp]
    \centering
        \scriptsize
     \begin{tabular}{p{7.5cm}}
     \toprule
          \textbf{Title}:  What states have yet to file a "Stay at home order"?\\
          \hline
           \multirow{2}{7.5cm}{\textbf{OP Post}: Curious to know what states have still not done stay at home orders or cancelled all group activities over a few hundred people?}\\ \\
          \hline
          \multirow{2}{7.5cm}{\textbf{Reply 1}: Texas is still insisting on being open because the small barely tested towns have few confirmed cases. All the big cities are on lock-down though.}\\ \\ \\
          \hline
         \textbf{Reply 2}: Missouri Governor refuses to but most cities have issued their own orders.\\
          \hline
          \multirow{4}{7.5cm}{\textbf{Reply 3}: Arkansas Governor says we only have two small hotspots and social distancing is working for the rest of the state so the only lockdowns are local and not state wide. But only essential services have been open for weeks and they have closed all parks. Some towns have established curfews}\\ \\ \\ \\
          \hline
         \textbf{Reply 4}: South Carolina is still wide open. Our governor is a douche. \\
        \hline
          \multirow{2}{7.5cm}{\textbf{Reply 5}: Oklahoma governor won’t do it for the whole state so the city mayors have started doing it themselves.} \\ \\
          \hline
         \textbf{Reply 6}: Missouri has not yet.\\
          \bottomrule
     \end{tabular}
     \caption{Example of Reddit thread 2}
    \label{tab:my_label}
 \end{table}
\normalsize
 \noindent \textbf{Summary of main story:} Users discuss the measures taken against the spread of coronavirus in their own states. Decisions of state governors are questioned in comparison to other cities and towns.
 \noindent \textbf{Reasoning:} The thread contains mostly factual information and few opinions (Only Reply 4 openly discusses user reaction), that is why only the main story of the thread is summarised.

\subsection{Human Evaluation Analysis} \label{apen-HEA}

The poor performance of WassOS on Amazon on referential clarity in Table \ref{HE_all} determines us to consider whether this is due to the difference in data. For this reason,
we investigated the poor performance of WassOS on Amazon with respect to referential clarity by counting the respective number of pronouns on Amazon and Twitter.
For each group/cluster in Amazon/Twitter the minimum number of reviews/tweets is 10. Therefore, we randomly selected 2000 groups/clusters from the datasets
, then we randomly selected 10 reviews/tweets from each group/cluster and counted the total number of pronouns. We repeated this process 10 times and averaged the final results. 76442.2 pronouns were obtained for Amazon as opposed to 21371.1 for Twitter. This confirms  that the referential relationship in Twitter is relatively simple compared to Amazon.

\subsection{Examples of summaries generated in different ablation settings } \label{apen-example-ablation-summaries}

\begin{table*}[!htbp]
	\label{summary result}
	\begin{center}
    \centering

	\begin{tabular}{|p{22mm}|p{115mm}|}

			\hline 
  
		Gold      & When I ordered this, I didn't know what to expect. I'm pleasantly surprised. It's plastic, but very convenient and the unit fits very well into my Zippo case. You can fine tune your preference as the torch adjusts very nicely. It works great. I also had issues with closing the cap of Zippo. \\
		\hline
		WassOS    &     This thing is great for the zippo case. It is very easy to use, easy to clean, and easy to keep clean. The only drawback is that it's a little hard to get the cap off, but that's a minor issue.
 \\
				\hline
		WassOS-dis         & I bought this for my zippo, and it works great.The only thing i don't like about it is that it doesn't come with a cap, but it does the job.
\\
		\hline
		WassOS-trans         &
		I ordered the head and square arm to get a square cup to water.I then it is my jam and they are mad so they start to jam it down the street department.This thing is the zippo alot better
		 \\ 
				\hline
			\end{tabular}
	\end{center}
	\caption{\label{ablation-amazon} Ablation experiment, Amazon summaries of ablation models}
\end{table*}

\begin{table*}[!htbp]
	\label{summary result}
	\begin{center}
    
			\begin{tabular}{|p{22mm}|p{115mm}|}
			
			\hline 
  
		Gold      & majority only just of tweets thank carers for their huge contribution. a second large subject discusses support for carers.  \\
		\hline
		WassOS    &     carersweek and we want to thank carers across the uk, you make a huge contribution to families  communities!
 \\
				\hline
		WassOS-dis        & Free events for carers in the uk - thank you for the pledge we want to thank our carersweek
\\
		\hline
		WassOS-trans          &
		See the carersweek time to celebrate carers and celebrate their skills and their isolated.Carers make to fix x.
		 \\ 
				\hline
			\end{tabular}
	\end{center}
	\caption{\label{ablation-twitter} Ablation experiment, Twitter summaries of ablation models}
\end{table*}

\subsection{Examples of Twitter opinion clusters and summaries generated}
\label{apen-examples-twitter}

\begin{table*}[!htbp]

	\begin{center}
    
			\begin{tabular}{|p{15mm}|p{110mm}|}
			
			\hline 
  
		Gold      & majority of tweets salute Manchester united and England footballer \textcolor[RGB]{18,148,168}{Rio Ferdinand as he retires}, \textcolor[RGB]{148,18,168}{some pointing to his future}, \textcolor[RGB]{228,148,28}{others recalling his early days at west ham}. \\
		\hline
		Copycat         & \textcolor[RGB]{18,148,168}{Good luck on your retirement Rio Rerdinand}, good luck with your services to see you doing a great job
 \\
				\hline
		Coop         & \textcolor[RGB]{18,148,168}{Good luck to your retirement} Rio Ferdinand's wife of the career to see you.
\\
		\hline
		WassOS          &
		\textcolor[RGB]{18,148,168}{Good luck on your retirement Rio Ferdinand}. \textcolor[RGB]{148,18,168}{Good luck with the future!} Hopefully see you doing some punditry.
		 \\  \midrule[1pt]

		Tweet1      & Good luck @rioferdy5 with retirement from football, and all the best in future endeavours.
		\\
		\hline
		Tweet2         & @rioferdy5 good luck Rio in your retirement from football x
 \\
				\hline
		Tweet3         & Happy retirement @rioferdy5 good luck with the future!  RioFerdinand.
\\
		\hline
		Tweet4          &	@rioferdy5 good luck with retirement Rio, thanks for your services to football and the national team. See u on our screens soon i'm sure.
		 \\
		 \hline
		 Tweet5       &    Be remiss not to wish @rioferdy5 a happy retirement and good luck! Off to the  for the FACupFinal.
		 \\
				\hline
		Tweet6       &    Good luck on your retirement Rio Ferdinand,loyal servant un your westham days \& true local boy,wish you well for the future, keep going fella.
		 \\
				\hline
		Tweet7       &   @rioferdy5 good luck with retirement Rio, thanks for your services to football and the national team. See u on our screens soon i'm sure.@rioferdy5 a happy retirement and good luck! Off to the  for the FACupFinal.
		 \\
				\hline
		Tweet8       &  @rioferdy5 Absolute legend. Good luck in retirement Rio! Hopefully see you doing some punditry.
		 \\
				\hline
			\end{tabular}
			\caption{Twitter summaries provided by three different models. We highlight information overlap between the gold standard and the generated summaries. The result shows that WassOS achieves the best performance for meaning preservation. } \label{tweet_summary result}
	\end{center}
\end{table*}

\subsection{Examples of Amazon reviews and corresponding summaries generated}
\label{apen-examples-review-summaries}

\begin{table*}[!htbp]

	\begin{center}
    
			\begin{tabular}{|p{15mm}|p{110mm}|}
			
			\hline 
  
		Gold      & This lived up to expectations after the product was featured on Shark Tank. \textcolor[RGB]{148,18,168}{The magnet is very strong, and will work anywhere on clothing. Glasses won't slip or fall.} However, those with pacemakers should avoid due to the strong magnet. Overall, recommend. \\
		\hline
		Copycat         & This is a great product for the price. I bought it for my daughter and she loves it. Only issue is that when you open the box, you have to turn off the glasses, but once you get used to it.
 \\
				\hline
		Coop         & This is a great product. It's very easy to use and the price is right. The only thing is that it doesn't have a strong magnet so you can take it off when you're on the go. It's just a little too small for your needs.
\\
		\hline
		WassOS          &
		This is easy to use and easy to clean. \textcolor[RGB]{148,18,168}{The magnet is strong enough to hold the glasses securely.} It is very easy to put on and take off. It has a strong magnet that holds the glasses in place.
		 \\  \midrule[1pt]

		Rev1      & what else can be said..... this product is revolutionary. i cannot believe something so simple could be so effective and i wonder why i didn't see this myself! kudos to the inventor. this magnet is strong and putting your glasses through the space while working is easy and just as easy to retrieve them when needed. necessity is the mother of invention....
		\\
		\hline
		Rev2         & these little guys are amazing,, the magnetic is very strong,, your glasses will not fall out when bending over.. yeah maybe a crystal does fall out so what,, just glue it back in and it stays,, you'll always know where your glasses are,, love mine.
 \\
				\hline
		Rev3         & ReadeRest Eye-Glass Holder-Magnet broke off after one week and I have no phone number to call the company. For the week I had it, it worked fine on my shirt but one of the front round magnets separted from the glass holding piece. A piece of junk!
\\
		\hline
		Rev4          &	I wish I could use this, but my cardilogist said no.... even though I have a fairly new, high-fangled pacemaker. He doesn't have any problem with me having a cell phone in my shirt pocket, but said this magnet is just too powerful.
		 \\
		 \hline
		 Rev5       &    i saw this product on shark tank and wanted it then. when i ordered it and tried it i was very impressed. the magnet is super strong and you can place it anywhere on your clothing. i bought 2 and will order more for christmas. 
		 \\
				\hline
		Rev6       &    fantastic product! ive had to order more now that ive shown to everyone. they all want one. can be placed anywhere on your clothing and blends right in. had people think it was part of the clothing. great gift / stocking stuffer.
		 \\
				\hline
		Rev7       &  great product! very strong magnet that works in holding and keeping your glasses secure. well worth the price that i paid for it... first showing of this product was on 'shark tank' and it has lived up to expectations!
		 \\
				\hline
		Rev8       &  i like it, keeps the glasses handy but when you bend over, they can fall out. so good idea, but has its limits. perhaps if it didn't stick out as far, that would hold the glasses tighter.
		 \\
				\hline
			\end{tabular}
			\caption{Amazon summaries provided by three different models. We highlight information overlap between the gold standard and the generated summaries.} \label{amazon_summary result}
	\end{center}
\end{table*}

\begin{table*}[!htbp]

	\begin{center}
    
			\begin{tabular}{|p{15mm}|p{110mm}|}
			
			\hline 
  
		Gold      & A discussion around having a sore arm following a Covid 19 vaccination. People share their stories of discomfort from different vaccines and ideas on how to ease the pain. 
		\textcolor[RGB]{148,18,168}{There is agreement that a sore arm is a very common side effect of the vaccine, and that the problem should pass in two or three days.}
		\\
		\hline
		Copycat         & I don't think that the first two shots would have to be able to comply with my first two years.
 \\
				\hline
		Coop         & the cdc is two very minimal, and its not the china virus. it's a few arm so much.
\\
		\hline
		WassOS          &
		The nurse said it was a half dose of the first two shots and \textcolor[RGB]{148,18,168}{my first two and was a sore arm after 2 days.}
		 \\  \midrule[1pt]

		Reddit1      & I got the booster yesterday. The nurse said it was a half dose. Today my arm aches like mad. Is this common? I don't recall the first two feeling like this. Injection site pain
		\\
		\hline
		Reddit2         & An achy arm is probably the most common COVID vax side effect of them all. I’ve had an achy arm after all 3 of my shots.
 \\
				\hline
		Reddit3         & My boostered arm ached more than doses one and two. It was gone within 3 days. The ache, not the arm.
\\
		\hline
		Reddit4          &	I definitely had a sore arm after my booster. Ibuprofen and hot tub helped me cope with it.
		 \\
		 \hline
		 Reddit5       &    I got Pfizer and my arm hurt way more from my booster than my first two shots and it was also very itchy. It probably lasted three days. I took an Epsom salt bath for the pain and I think that it really helped!
		 \\
				\hline
		Reddit6       &    I had AZ for my first two and was pretty much side effect free, I got a Moderna booster and my arm hurts so much, it's all red around the injection side and its still swollen and painful after 2 days.
		 \\
				\hline
		Reddit7       &   Totally normal. Inconvenient, but normal. Take some tylenol or ibuprofen if you haven't, that should help.
		 \\
				\hline
		Reddit8       &  When I got my booster, I was really tired the next day, but my arm also was REALLY sore for 2 days afterwards, to the point I could barely lift it above my shoulder. That also did not happen with my first 2 shots. Should clear up after a few days and you will be right as rain with a booster to boot!
		 \\
				\hline
			\end{tabular}
			\caption{Reddit summaries provided by three different models. We test the Reddit threads directly using the pre-trained parameters on Twitter. We highlight information overlap between the gold standard and the generated summaries.} \label{tweet_summary result}
	\end{center}
\end{table*}